\documentclass[10pt,twocolumn,letterpaper]{article}

\usepackage[final]{cvpr}      

\usepackage{graphicx}
\usepackage{amsmath}
\usepackage{amssymb}
\usepackage{booktabs}

\usepackage{multirow}
\usepackage{times}
\usepackage{epsfig}
\usepackage{amssymb}
\usepackage[table]{xcolor}
\def\arrvline{\hfil\kern\arraycolsep\vline\kern-\arraycolsep\hfilneg}
\usepackage{etoolbox}

\newcommand\blfootnote[1]{%
	\begingroup
	\renewcommand\thefootnote{}\footnote{#1}%
	\addtocounter{footnote}{-1}%
	\endgroup
}

\usepackage[pagebackref,breaklinks,colorlinks]{hyperref}

\usepackage[capitalize]{cleveref}
\crefname{section}{Sec.}{Secs.}
\Crefname{section}{Section}{Sections}
\Crefname{table}{Table}{Tables}
\crefname{table}{Tab.}{Tabs.}

\def\cvprPaperID{3} 
\def\confName{CVPR}
\def\confYear{2022}

\begin{document}

\title{CONSENT: Context Sensitive Transformer for Bold Words Classification}

\author{Ionut-Catalin Sandu \hspace{3mm}  Daniel Voinea \hspace{3mm}  Alin-Ionut Popa \\{\tt\small \{sanion, dvoinea, popaaln\}@amazon.com} \\ 
	Amazon Inc.\\}
\maketitle

\begin{abstract}
We present \textbf{CONSENT}, a simple yet effective \textbf{CON}text \textbf{SEN}sitive \textbf{T}ransformer framework for context-dependent object classification within a fully-trainable end-to-end deep learning pipeline. We exemplify the proposed framework on the task of bold words detection proving state-of-the-art results. Given an image containing text of unknown font-types (\eg Arial, Calibri, Helvetica), unknown language, taken under various degrees of illumination, angle distortion and scale variation, we extract all the words and learn a context-dependent binary classification (\ie bold versus non-bold) using an end-to-end transformer-based neural network ensemble. To prove the extensibility of our framework, we demonstrate competitive results against state-of-the-art for the game of rock-paper-scissors by training the model to determine the winner given a sequence with $2$ pictures depicting hand poses.

\end{abstract}

\begin{figure} [!htbp]
	\begin{center}
		\begin{tabular}{cc}
			\includegraphics[width=0.49\linewidth]{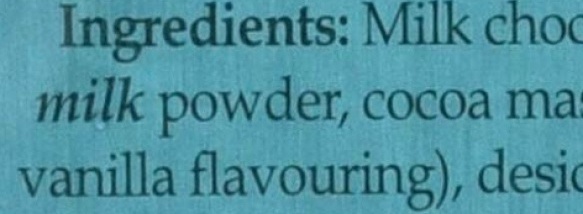}
			\includegraphics[width=0.46\linewidth]{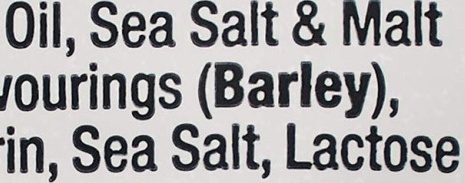} 
		\end{tabular}
	\end{center}
	\caption{\textbf{Image examples with context dependent bold-style words}. It is distinguishable in the above images, that the notion of bold-style word is dependent to the image context (word \textit{ingredients} in the left image w.r.t. to the normal words in the right image).}
	\label{fig:bold_samples}
\end{figure}

Object localization \blfootnote{ \href{https://sites.google.com/view/fgvc9/home}{$9$\textsuperscript{th} Workshop on Fine-Grained Visual Categorization}, CVPR $2022$, New Orleans, Louisiana.} and detection \cite{ren2015faster, He_2017_ICCV, redmon2016you, Lin_2017_ICCV, tan2020efficientdet} are some of the most well-known and intensely approached problems from the computer vision literature. Historically, for these types of tasks the main focus is on discovering discriminative features for the searched objects with respect to the image space where the objects of interest are localized. This tends to become problematic when the task at hand requires analysing the surrounding image context, prior to taking a decision regarding the object type. One such scenario is the problem of bold font-style detection. Given an image containing text, determine which words are bold and which are not. It is necessary for such an assessment to be performed locally (\ie at image level), and not globally (\ie across different images). A visual illustration of this issue can be seen in figure \ref{fig:bold_samples}.   

This is by no means a simple problem as there is the local image context to incorporate in the statistical inference and also there is a multitude of factors which contribute to its complexity such as unknown font-type(s) used inside the image, unknown text language (no prior information regarding the visual structure of the text characters), different foreground-background colors or intensity variations, multiple degrees of freedom with respect to camera viewpoint, skewness or just random affine noise.

\begin{figure*}[t]
	\begin{center}
		\includegraphics[width=0.85\linewidth]{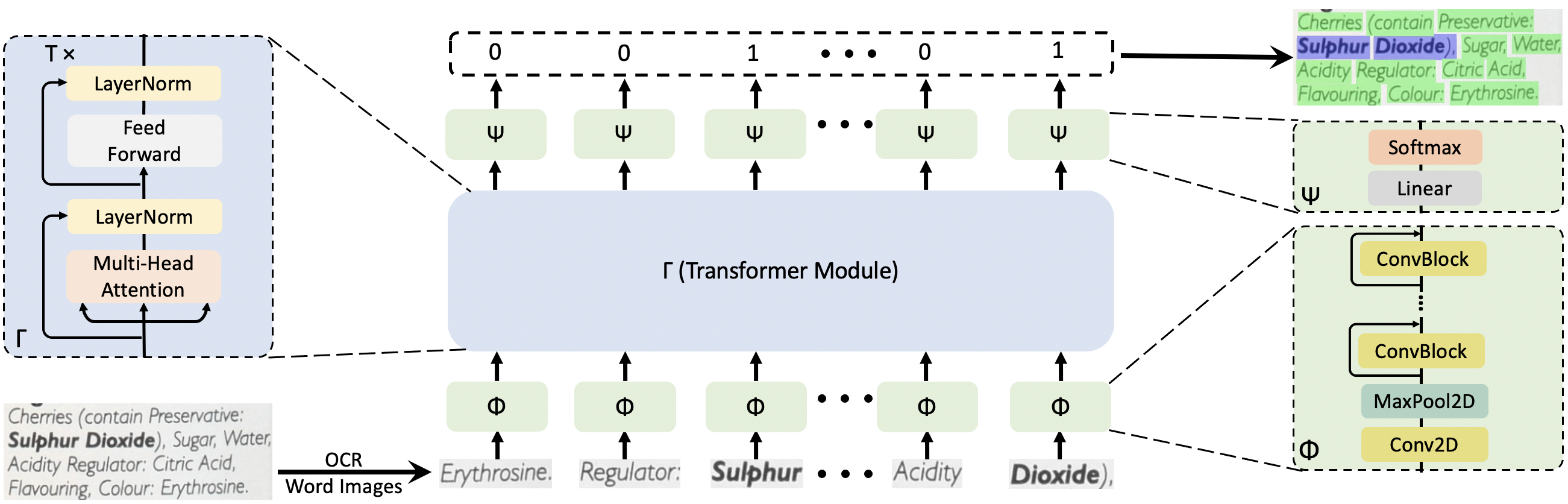}
	\end{center}
	\caption{\textbf{Detailed view of CONSENT framework.} Given an image containing text, we first extract the cropped images containing words using an OCR engine \cite{textract}. Next, these are passed through a feature extractor $\Phi$ as a sequence to obtain specialized features for the bold word detection problem. The obtained features are then passed through a self-attention embedder (transformer) module $\Gamma$ which learns an implicit statistic between the elements of the sequence. Finally, these go through a decoding module $\Psi$ to obtain the final class estimates for each word-level image patch. Each component (\ie $\Gamma$, $\Phi$, $\Psi$)is emphasized using a magnifying zoom view.}
	\label{fig:detailed_overview}
\end{figure*}

Our pipeline is summarized with the following computational steps: given an image containing text, (a) we identify all the words by using a state-of-the-art optical character recognition (OCR) system, (b) group the extracted word-based image regions into sequences and pass them through our proposed \textbf{CONSENT} module. \textbf{CONSENT} consists of an image-based feature extractor backbone, followed by an encoding reasoning system via a self-attention embedder \cite{vaswani_nips_2017} and a feed-forward network head which acts as a classifier for the word type.  The motivation behind this is to allow a self-attention embedder (\textit{i.e.} transformer) to pay attention to all words from a sequence in order to specialize the backbone network in extracting relevant features for the bold versus non-bold comparison. An in-depth view is illustrated in figure \ref{fig:detailed_overview}. 

\section{\textbf{Related Work}}

There is a remarkable line of work dedicated to OCR \cite{tesseract, textract, Wigington_2018_ECCV}, however most of them focus on the topic of text localization and understanding. There are some \cite{zhu2001font, bychkov2020using, moussa2010new, pengcheng2017chinese,wang2015deepfont,oh2016smooth,tensmeyer_icdar_2017} dedicated to understanding text style related features. They are mostly focused on font-type comprehension which is globally discriminative (\ie at dataset level).  For example, let us go over the method of \cite{wang2015deepfont}. The authors propose a multi-class image classification convolutional neural network which for a given text image patch predicts the font-type of the text (\eg Serif, Script, Arial). However, this does not give any relevant information with respect to the font-style of the text (\eg italic, bold, caps lock, underline). The work of \cite{tesseract} claims to have such a feature, however, when testing their API it is unavailable. 

Modern networks employ transformers-like \cite{vaswani_nips_2017,dosovitskiy2020,bao2021beit,xie2021self,jaegle2021perceiver,tay2021charformer,dai2019transformer} architectures that accounts for information from different representation states, thus eliminating the bottle-neck effect caused by the usual encoding-decoding schemes. One such example is \cite{dosovitskiy2020} where the authors adapt the transformer model for the task of image classification, by doing an image-patch level embedding combined with positional encoding of the patch with respect to the image grid. 
In \cite{hu2018relation}, the authors make use of the attention mechanism to act as a geometric weight for capturing the spatial relationship between object instances. Our method is different in the sense that we leverage the transformer architecture to build context-sensitive statistics of the target classes (\ie bold vs. non-bold). 

\section{\textbf{Dataset and Annotation Protocol}} \label{sec:dataset}

\paragraph{\textbf{BONN Dataset}}

We collected images from the public web, in particular images of food products. We covered multiple languages and as many subcategories as possible, to have a high diversity in terms of text descriptions, font-styles and font-types.  
We successfully downloaded a total of $11,526$ images with ingredients or nutritional descriptions. Given the current progress in the OCR field, we leveraged an existing state-of-the-art OCR engine \cite{textract} for the extraction of word-level image patches for the $11,526$ images. After removing those with low resolution and poor OCR detections, we ended up with $9,232$. Lastly, we obtained a total of $294,197$ word-level patches. We will refer to this data collection as \textbf{BONN} Dataset (\textbf{BO}ld w\textbf{O}rds in i\textbf{N}gredie\textbf{N}ts).

\paragraph{\textbf{Annotation Protocol and Dataset Statistics}}

In order to add bold and non-bold word annotations, we create a JavaScript based plugin, where we display each individual image with bounding boxes plotted on top of each extracted word-level image patch. Prior to starting the annotations process, we observed that there is higher non-bold word ratio at image level, so we made all the word types in the plugin interface as non-bold and we asked the annotators to change the word type into bold accordingly. After this annotation process the initial pool of $294,197$ words got split into $267,773$ non-bold words and $26,424$ bold words. More details regarding the dataset statistics are in table \ref{tbl:extra_details} - \textit{(Right)}. It is noticeable that there is a clear data imbalance towards the non-bold type. The dataset contains images with a couple of words (minimum of $1$) and images with hundreds of words (maximum of $701$ words). 

\section{\textbf{Methodology}}

Given an image $\mathbf{I} \in \mathbb{R}^{h \times w \times 3}$ with text, we want to classify all the word-level image patches as either bold or non-bold.
After applying \cite{textract}, we obtain a pool of $N$ word-level image patches denoted by $\mathbf{P} = \{P_i \; | \; P_i \in \mathbb{R}^{h_i \times w_i \times 3}, \; i = 1 \dots  N\}$, where $N$ is the number of words from image $\mathbf{I}$. For each $P_i$ there is a corresponding binary label attached, marking it as either bold or non-bold. We will denote the attached array of labels with $\mathbf{L} = \{0, 1\}^N$. Thus, our train samples consists of pairs $(\mathbf{P}_j, \mathbf{L}_j)$ where $j = 1 \dots N_{images}$ and $N_{images}$ being the total number of data samples. For ease of notation and understanding, we will refer to $\mathbf{P}_j$ and $\mathbf{L}_j$ as $\mathbf{P}$ and $\mathbf{L}$ respectively. 

\paragraph{\textbf{Letter Morphology Voting}} \label{sec:voting}

The first noticeable thing is the fact that the letter should be thicker when comparing bold with non-bold. Thus, we apply hand-crafted image processing heuristics to recover the bold words.  

\begin{figure} [!htbp]
	\begin{center}
		\begin{tabular}{ccc}
			\includegraphics[width=0.28\linewidth]{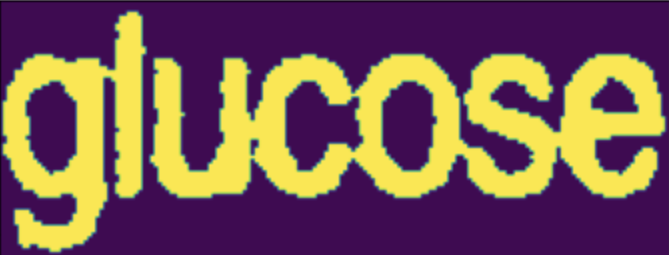}
			\includegraphics[width=0.28\linewidth]{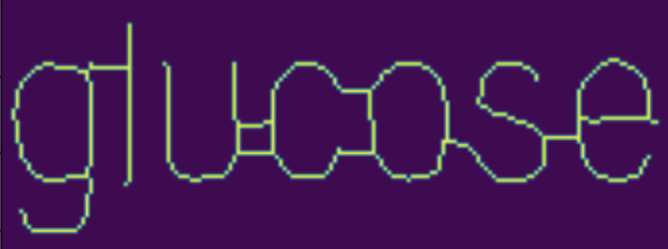}
			\includegraphics[width=0.28\linewidth]{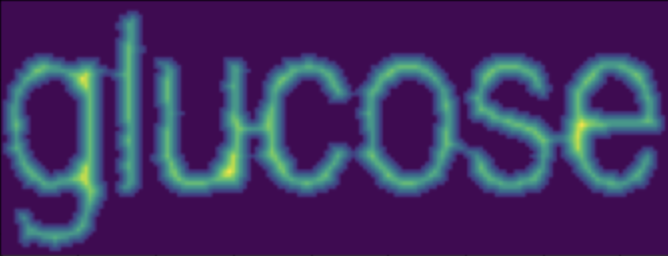} 
		\end{tabular}
	\end{center}
	\vspace{-3mm}
	\caption{\textbf{Letter-Level Morphology Operations}. From left to right we have letter-level segmentation mask $S_i$, skeletonization mask $S^{skel}_i$ and foreground distance transform mask $S^{dist}_i$.}
	\label{fig:letter_morphology}
\end{figure}

For each word-level image patch $P_i \in \mathbf{P}$, we apply image binarization to obtain the letter-level segmentation mask, denoted by $S_i$ with $i = 1 \dots  N$. 
For each mask $S_i$, we apply two morphological based operations: (1) skeletonization \cite{ersoy2011skeleton} denoted by $S^{skel}_i$ and (2) foreground mask distance transform denoted by $S^{dist}_i$ (see figure \ref{fig:letter_morphology}). The intuition is that $S^{skel}_i$ should give us the inner most pixels with respect to the letter shape and $S^{dist}_i$ gives us the distance towards the closest edge of the letter. 

As a result, for each word, we have	 $\theta(P_i) = \{S^{dist}_i(x, y) \; | \; S^{skel}_i(x, y) = 1  ,  i = 1 \dots  N\} $. Basically, $\theta(P_i)$ gives us a thickness measurement computed over all skeleton pixels of a word. By iterating over each $\theta(P_i)$, we obtain a letter thickness measurement for $\mathbf{I}$ in the form of $\Theta_\mathbf{I} = \{ \theta(P_i) \; | \; i = 1 \dots N\}$.

Having $\Theta_\mathbf{I}$, and the prior knowledge that $\approx 10\%$ of the words are bold, we use of the following voting scheme to determine if a word is bold or not,

\begin{equation*}
	B(P_i) = \left\{
	\begin{array}{l l}
		1 & \quad \mbox{if $\mu(\theta(P_i)) > med(\Theta_\mathbf{I}) + \alpha \cdot \sigma(\Theta_\mathbf{I}) $} \\
		0 & \quad \mbox{if $\mu(\theta(P_i)) \leq med(\Theta_\mathbf{I}) + \alpha \cdot \sigma(\Theta_\mathbf{I}) $} \\
	\end{array} \right.
\end{equation*}

\noindent where $\mu(\cdot)$, $\sigma(\cdot)$ and $med(\cdot)$ represent the mean, the standard deviation and the median function, respectively. Parameter $\alpha$ is a voting threshold requiring validation.

\paragraph{\textbf{CONSENT: Context Sensitive Transformer}} \label{sec:consent}

Given the previously defined sequence $\mathbf{P}$ of word patches from an image $\mathbf{I}$, a model should learn their corresponding labelling $\mathbf{L}$ by performing an inference constrained by the local image context. To accomplish this particular task, our proposed model, \textbf{CONSENT}, incorporates three distinctive components (\textbf{I}, \textbf{II} and \textbf{III}):

\noindent \textbf{(I)}, $\mathbf{\Phi}$, a feature extractor backbone operating at sub-image level. It maps a patch of words $\mathbf{P} = \{P_i \; | \; i = 1,\dots,N\} $ to a sequence of embeddings $\mathbf{E} = \{E_{i} \; | \; i = 1, \dots, N\}$. For this purpose, we got inspiration from the pretrained classifiers \cite{He2015, howard2017mobilenets, iandola2016squeezenet, Huang_2017_CVPR} on ImageNet \cite{deng2009imagenet}. The intuition is to look at word-level image crops and construct embeddings discriminative enough for the task of bold word detection. Thus, we have $\mathbf{E} = \mathbf{\Phi}(\mathbf{P})$.

\noindent \textbf{(II)}, $\mathbf{\Gamma}$, a self-attention embedder that receives as input a sequence of embeddings $\mathbf{E}$ and outputs another sequence $\hat{\mathbf{E}}_t = \{\hat E_{i}^t \; | \;  i  = 1\dots N\}$ as a result of applying attention concerning their class type. It consists of $t  = 1\dots T$ stacked encoding layers, where each encoding layer consists of multi-head attention, layer normalization \cite{ba2016layer} and feed-forward together with residual-block connections. The motivation is to let the self-attention embedder learn an implicit statistic between the sequence elements and specialize the feature representations $\hat{\mathbf{E}}_t$ with respect to binary classification given the local context. Given that $\hat{\mathbf{E}}_0 = \mathbf{E}$, we have $\hat{\mathbf{E}}_t = \mathbf{\Gamma}(\hat{\mathbf{E}}_{t-1})$.

\noindent \textbf{(III)},$\mathbf{\Psi}$, a final decoding step, which maps the output of the self-attention embedder to our binary representation. It consists of a fully connected layer followed by a \texttt{softmax} layer that maps each element of $\hat{\mathbf{E}}_T$ to a two dimensional vector $\hat{\mathbf{L}} = \{0, 1\}^N$, representing probabilities for each class (non-bold and bold respectively). Lastly, we have the formalization $\mathbf{CONSENT(P)} = \mathbf{\Psi}(\hat{\mathbf{E}}_T) = \hat{\mathbf{L}}$.

Having all the components defined, we use the binary cross-entropy loss function to optimise the weights of the model.

\begin{table*}
	\scalebox{0.68}{
		\begin{tabular}{|*{6}{c|}}
			\cline{3-6}
			\multicolumn{2}{c|}{} & \multicolumn{4}{c|}{\textbf{Embedding Size}} \\
			\cline{3-6}
			\multicolumn{2}{c|}{} & \multicolumn{1}{c|}{$\mathbf{32}$} & \multicolumn{1}{c|}{$\mathbf{64}$} & \multicolumn{1}{c|}{$\mathbf{128}$} & \multicolumn{1}{c|}{$\mathbf{256}$} \rule{0pt}{8pt}\\ 
			\hline 
			\rule{0pt}{8pt}	\multirow{3}{*}{\rotatebox[origin=c]{90}{\textbf{\# of Stacks}}} &  $\mathbf{2}$ & \cellcolor{orange!76} 0.88 & \cellcolor{orange!62} 0.85 & \cellcolor{orange!76} 0.88& \cellcolor{orange!66} 0.86 \\  
			\cline{2-6}
			\rule{0pt}{8pt} & $\mathbf{3}$ & \cellcolor{orange!62} 0.85 & \cellcolor{orange!80} 0.89 & \cellcolor{orange!80} 0.89 & \cellcolor{orange!71}0.87\\ 
			\cline{2-6}
			\rule{0pt}{8pt} & $\mathbf{4}$ & \cellcolor{orange!80} 0.89 & \cellcolor{orange!95} \textbf{0.91}   & \cellcolor{orange!71} 0.87 & \cellcolor{orange!80} 0.89\\ 
			\cline{2-6}
			\rule{0pt}{8pt} & $\mathbf{5}$  & \cellcolor{orange!90} 0.9 & \cellcolor{orange!80} 0.89  & \cellcolor{orange!66} 0.86 & \cellcolor{orange!40} 0.76\\ 
			\hline
	\end{tabular}}
	\hfill
	\scalebox{0.68}{
		\begin{tabular}{|l|c|c|c|c|}
			\hline
			\textbf{Method} & \textbf{Precision} & \textbf{Recall} & \textbf{F1-Score} & \textbf{Image Acc.(\%)} \\
			\hline\hline
			Morphology Voting & $0.71$ & $0.73$ & $0.72$ & $40$ \\
			\hline
			DenseNet \cite{Huang_2017_CVPR} & $0.8$ & $0.47$ & $0.59$ & $38$ \\
			\hline
			CONSENT w/o \cite{Lin_2017_ICCV} & $0.91$ & $0.88$ & $0.89$  & $75$\\
			\hline
			\textbf{CONSENT w \cite{Lin_2017_ICCV}} & $\textbf{0.92}$ & $\textbf{0.88}$ & $\textbf{0.90}$ & $\textbf{78}$ \\
			\hline
	\end{tabular}}
	\hfill
	\scalebox{0.68}{
		\begin{tabular}{|l|c|c|c|}
			\hline
			\rule{0pt}{11pt} 	\textbf{Statistic Type} & \textbf{All} & \textbf{Bold} & \textbf{Non-Bold}\\
			\hline\hline
			\rule{0pt}{12pt} 	Min / Max & $1$ / $701$ & $0$ / $147$ & $1$ / $689$ \\
			\hline
			\rule{0pt}{12pt}  Mean $\pm$ std. & $32 \pm 42$ & $4 \pm 6$ & $29 \pm 38$ \\
			\hline
			\rule{0pt}{12pt} 	Median & $22$ & $2$ & $19$ \\
			\hline
	\end{tabular}}
	\caption{\textbf{\textit{(Left)}} We perform an ablation study on \textbf{BONN} validation set by varying the embeddings of the self-attention embedder $\Gamma$ as well as the number of encoding stacks to decide on the best performing hyperparameter configuration in terms of F1-score. \textbf{\textit{(Middle)}} \textbf{BONN Test Set Bold-Word Class Metrics}. We evaluate our proposed \textbf{CONSENT} model and the additional baselines: morphology based voting and DenseNet. \textit{\textbf{Bold-Word Class Results (columns $\mathbf{2}$, $\mathbf{3}$ \& $\mathbf{4}$):}} The morphology voting performs as good as the image quality allows. The best performance is obtained with \textbf{CONSENT}. Additionally, please notice that by using the focal loss \cite{Lin_2017_ICCV} we achive a slight improvement. \textit{\textbf{Image-Level Results (column $\mathbf{5}$)}} The image level performance is proportional with the bold-word class metrics, however at lower values as the accuracy criteria is much more rigid. \textbf{\textit{(Right)}} We aggregated image level statistics for all image-patch types for our proposed \textbf{BONN} dataset. There exists a high intra-class variation in terms of number of bold and non-bold words per image and a high non-bold class imbalance.}
	\label{tbl:extra_details}
\end{table*}

\begin{figure} [!htbp]
	\begin{center}
		\begin{tabular}{cc}
			\includegraphics[width=0.475\linewidth]{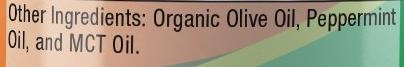}
			\includegraphics[width=0.475\linewidth]{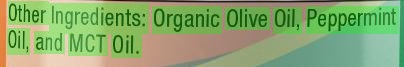} \\
			\includegraphics[width=0.475\linewidth]{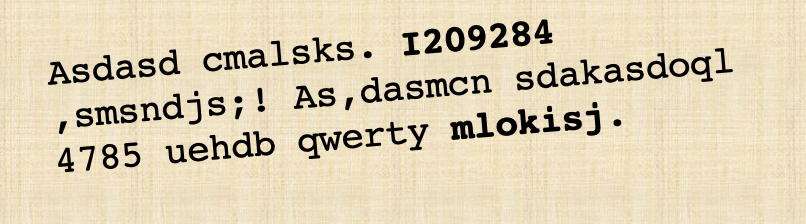}
			\includegraphics[width=0.475\linewidth]{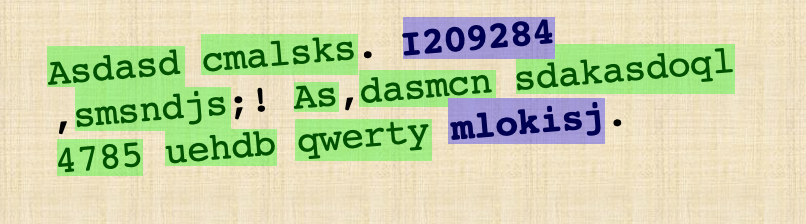} 
		\end{tabular}
	\end{center}
	\caption{\textbf{Sample classification results using CONSENT - blue for bold / green for non-bold.} The model is robust with respect to illumination changes and different foreground-background colors (row $1$) and semantic meaning of text (row $2$).}
	\label{fig:sample_results}
\end{figure}

\section{\textbf{Experiments}}

We conduct experiments on the proposed \textbf{BONN} dataset and the \textbf{Rock-Paper-Scissors} \cite{rps} dataset. We divided \textbf{BONN} in $3$ splits: train ($80\%$), test ($15\%$) and validation ($5\%$). It is worth mentioning that some of the images refer to the same product (\ie different viewpoint of the same object), thus, we might have the same product from different viewpoints. For a fair evaluation, we ensured all the images corresponding to a product are in the same split. 

Our main objective is to determine the bold words from the image. Thus, we report the metrics (\ie precision, recall and f1-score) only for the bold words class for each of our proposed approaches. Also, we look at macro performance level, by measuring the percentage of images where all the words are correctly predicted.

\begin{figure} [!htbp]
	\begin{center}
		\begin{tabular}{cc}
			\includegraphics[width=0.47\linewidth]{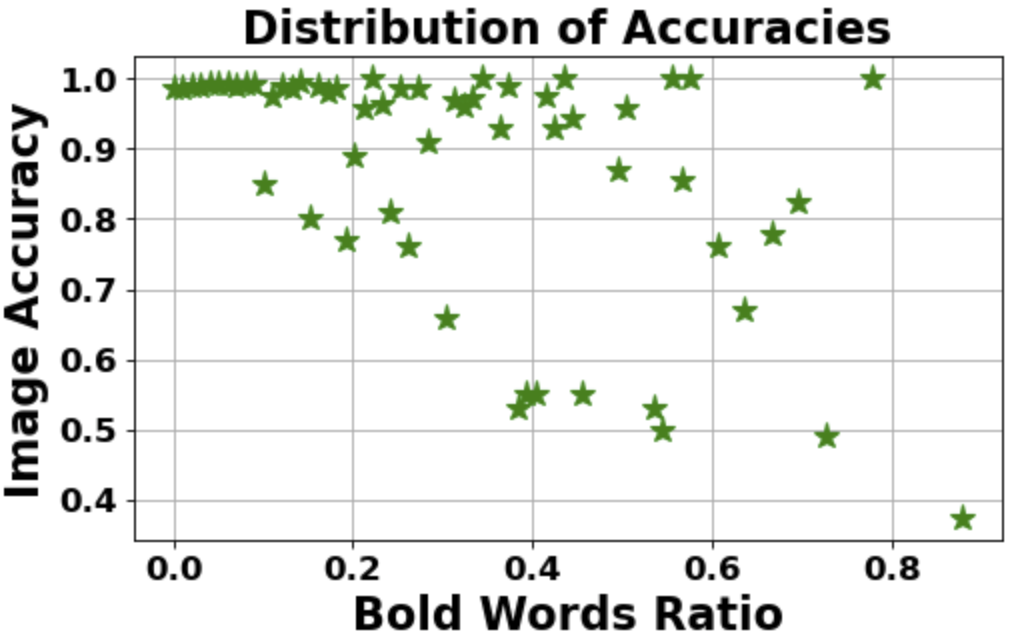}
			\includegraphics[width=0.474\linewidth]{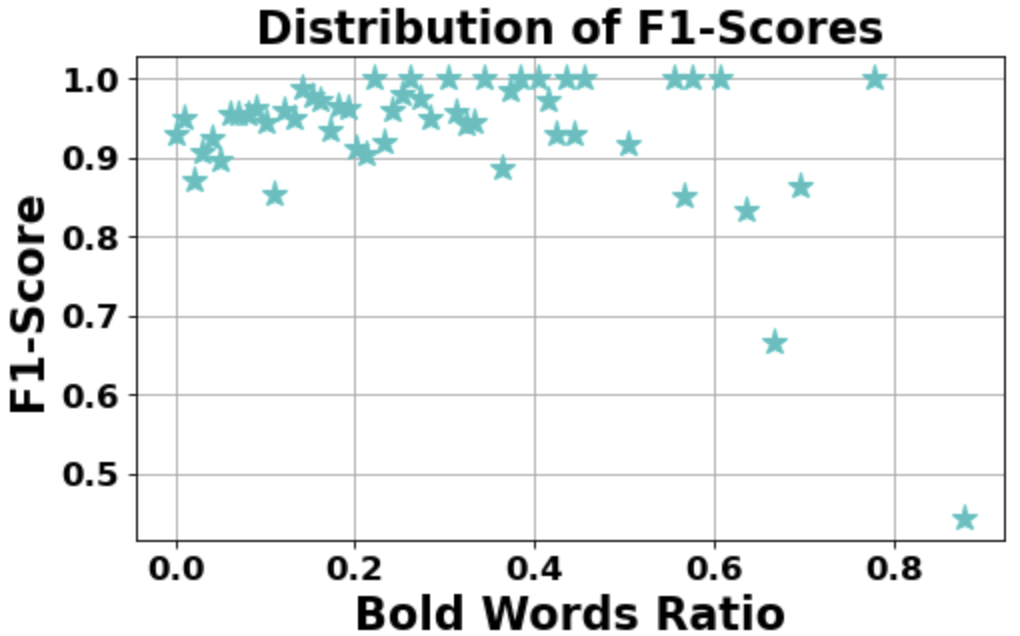} 
		\end{tabular}
	\end{center}
	\caption{\textbf{Distribution of performance w.r.t. bold words ratio.} We illustrate the performance of our model for image sets from \textbf{BONN} test set grouped according to their bold word ratios. Please notice that the performance is high for small ratios of bold words and as the ratio gets higher, the task gets more complex and challenging and the performance of the model slightly decreases.}
	\label{fig:performace_bold_ratio}
\end{figure}

\paragraph{\textbf{CONSENT}} The word-level image patches are split in blocks with fixed $3:4$ aspect ratio which are resized to $96 \times 128$. The best performance of the model was obtained by using sequence size of $100$, embedding size of $64$ and $T=4$ stacks of encoding layers. Performance details regarding the hyperparameter selection can be seen in table \ref{tbl:extra_details} - \textit{(Left)}. When selecting the image blocks for the sequence, we made sure that they were part of the same word-level image patches. We obtained an f1-score of $0.9$ (see table  \ref{tbl:extra_details} - \textit{(Middle)}). This further strengthens the argument that the bold word detection problem depends on the local context and prove that our approach can be successfully applied to this type of problems. Visual results of our model are illustrated in figure \ref{fig:sample_results}.  We also measure the performance at image level, see table \ref{tbl:extra_details} - \textit{(Middle)}, last column. For each image, we consider it as correct if all the words are correctly classified as bold or non-bold accordingly and incorrect otherwise. The best performance is obtained with the \textbf{CONSENT} model and the performance is at a peak of $78\%$ which emphasizes the complexity of the proposed dataset. In figure \ref{fig:performace_bold_ratio} we show the peformance of the model in terms of image accuracy corelated with the bold words ratio.

\paragraph{\textbf{Additional evaluation}} 
Lastly, we illustrate the versatility of the \textbf{CONSENT} pipeline. We tested our method on another context-dependent object classification task, the game of \textbf{Rock-Paper-Scissors} \cite{rps}. Traditionally, for this task, a multi-class image classifier is applied and the winner is decided based on the determined classes (\ie rock, paper or scissors). However, we use the initial data and generated image sequences with length of $2$. Instead of classifying each image individually, we want to decide the winner based on the context of the sequence. For example, for the sequences $[$\textit{rock}, \textit{scissors}$]$, $[$\textit{rock}, \textit{paper}$]$ and $[$\textit{rock}, \textit{rock}$]$,  the target scores are $[1,0]$, $[0,1]$ and $[0,0]$ respectively. We trained \textbf{CONSENT} on the generated setup and compaired against \texttt{ResNet18}. Using the test set of \cite{rps}, we generated $2,400$ individual games. The accuracy of our model is $\mathbf{94.4\%}$, compared with $\mathbf{94.7\%}$ obtained with a standard CNN \cite{He2015}, thus proving that \textbf{CONSENT} is able to achieve the same performance using a context-dependent problem formulation.

\section{\textbf{Conclusions}}

In this paper we present a principled, rigorous formulation for the problem of bold-word detection together with a novel and efficient solution based on transformers \cite{vaswani_nips_2017} and a new dataset, \textbf{BONN}, with $9,232$ images fully annotated with bold words. We compare our proposed method with two other strong baseline approaches, one based on letter morphology voting using letter thickness and a generic state-of-the-art image-classifier approach, proving we achieve superior results. We successfully applied \textbf{CONSENT} to the game of rock-paper-scissors, proving state-of-the-art results. In principle, with enough supervision, the pipeline can be applied to any generic object classification task which is sensitive with respect to the local context (\eg temporal segmentation of erroneous fitness repetitions, referee / player classification during a soccer match). 

{\small
\bibliographystyle{ieee_fullname}
\bibliography{consent_cvpr_22}
}

\end{document}